# Packaging and Sharing Machine Learning Models via the Acumos AI Open Platform

Shuai Zhao[1,2], Manoop Talasila[1], Guy Jacobson[1], Cristian Borcea[2], Syed Anwar Aftab[1], and John F Murray[1]

[1]AT&T Research Labs, Bedminster, NJ, USA
[2]New Jersey Institute of Technology, Newark, NJ, USA
sz255@njit.edu, talasila@research.att.com, guy@research.att.com
borcea@njit.edu, anwar@research.att.com, jfm@research.att.com

*Abstract*—Applying Machine Learning (ML) to business applications for automation usually faces difficulties when integrating diverse ML dependencies and services, mainly because of the lack of a common ML framework. In most cases, the ML models are developed for applications which are targeted for specific business domain use cases, leading to duplicated effort, and making reuse impossible. This paper presents Acumos, an open platform capable of packaging ML models into portable containerized microservices which can be easily shared via the platform's catalog, and can be integrated into various business applications. We present a case study of packaging sentiment analysis and classification ML models via the Acumos platform, permitting easy sharing with others. We demonstrate that the Acumos platform reduces the technical burden on application developers when applying machine learning models to their business applications. Furthermore, the platform allows the reuse of readily available ML microservices in various business domains.

## I. INTRODUCTION

In recent years, there has been tremendous excitement around and interest in the potential of ML technologies. Machine learning has been shown effective in solving a variety of practical problems such as disease detection [1], language translation [2], autonomous self-driving cars [3] and customer behavior prediction [4].

However, in practice, it is challenging to integrate ML models into application development environments [5]. Typically, ML models involve multi-stage, complex pipelines with procedures that are sequentially entangled and mixed together, such as pre-processing, feature extraction, data transformation, training, and validation. Improving an individual component may actually make the model accuracy worse if the remaining errors are more strongly correlated with the other components. Therefore, building models becomes a trial-and-error-based iterative process [6] which demands expert level knowledge in ML concepts to create and tune the ML models.

This requirement of technical ML know-how creates a heavy technical burden on traditional companies and small businesses, which may only have little or no technical expertise in machine learning [7]. In addition, training ML models can require computational resources with impracticable costs, such as GPUs or cluster computing environments. It is challenging for an average application developer to access affordable ML models to integrate into there applications.

To address that issue, we present a new ML packaging and distribution platform: Acumos. Acumos [8] was created to enable the packaging, cataloging, and sharing of ML models. It creates a separation between ML models, which are designed by machine learning experts and trained via a process of feedback from real-world results, and the surrounding infrastructure which deploys and runs the models. Acumos works like an app store which allows data scientists or modelers to upload their pre-trained models. Then Acumos packs the models and their dependencies into lightweight services. End users can download and deploy these services on the platform of their choice, either locally or in the cloud. The development of domain adaption [9], representation learning [10] and transfer learning [11] demonstrates the effectiveness of transferring pre-trained models to other similar datasets, whose distribution may be slightly different. The advantages of Acumos can be summarized as follows.

First, Acumos offers a one-stop convenient deployment service. Data scientists and modelers are more comfortable designing and testing the models locally. After they build and train their model, Acumos can freeze the model parameters and clone the whole running environment into a packed runnable service. The modelers are given complete freedom to build and train their models with their favorite tools and languages. As long as the languages are supported by Acumos, the modelers can on-board their models in our platform.

Second, Acumos offers model-level isolation. Real practices need to train multiple models over a single dataset for various tasks. For example, given a set of images, there may be multiple tasks, such as face detection, landmark detection,and mode detection. With Acumos, teams can work independently on different problems. Model-level isolation also facilitates the reuse and sharing of models with other similar applications without breaching model privacy.

Third, Acumos can help to distribute the robust and runable models from model experts to common end-users. In Acumos, we can treat ML models as black boxes which take well-defined inputs and generate output. End users do not require special knowledge on machine learning.

The remaining sections of this paper are organized as follows: Section II gives an overview of the background and related work on ML platforms; Section III explains the Acumos design and work processes; Section IV describes the ML architecture for Acumos as well as challenges and corresponding solutions; Section V presents a case study in using Acumos; Section VI discusses future work and limitation; and finally Section VII concludes the paper.

## II. RELATED WORK

There is a large number of packages and tools developed for machine learning. Some prominent and popular ones include scikit-learn [12], tensorflow [13], Spark MLlib [14], and rapidminer [15]. This paper is focused on the integration of ML models into application development and sharing models, rather than model development or improving model performance.

Compared to model development, integrating ML models into application development and sharing models have received relatively little attention in previous studies. Tensorflow-Serving [16] (TFS) encapsulates the production infrastructure of Google and serves TensorFlow-based models in production. Clipper [17] was developed concurrently with Tensorflow-Serving, and it shares similar goals and components with Tensorflow-Serving. The major difference is that Clipper was created and is maintained by research communities, and it is more general, also applying to non-Tensorflow models. But neither of the two platforms is able to share models with other users. Those platforms were designed only to deploy ML models in production environments.

Acumos integrates the functions of model packaging and sharing together in a single platform. Another system that allows the sharing of models is ModelDB [18], where models can be archived and accessed via source code, specifically those in the area of computational neuroscience. To some extent, OpenML [19] is a social sharing cloud for ML experts to share raw models. To actually use these raw models, ML experts have to go through the tedious task of configuring the model's specific environment manually. Compared to ModelDB and OpenML, Acumos offers easy-to-deliver end-to-end runnable services by packaging the model into a Docker image, which can be executed without any environment restrictions. Furthermore, Let us also note that Kubeflow[1], an ongoing project to deploy models into Kubernetes containers, could be integrated into Acumos in order to complement the containerization features of Acumos.

Major internet companies recently have begun offering machine learning as a service on their platforms, including Google's Cloud ML Engine[2] (previously called Prediction API), Amazon's AWS ML[3], and Microsoft's Azure ML[4]. Those platforms provide APIs to upload data and train models. Unfortunately, the designs and implementation specifications of these products are not publicly available [7]. One common limitation of these platforms is that they are restricted to their own particular proprietary cloud service and specific ML libraries, due to their commercial nature. For example, users are required to use Google Cloud as well as Tensorflow for Google Cloud ML Engine. Thus, they create a technical burden for data scientists to transfer and share their models with application developers, who may use other cloud providers or their own hardware. Another problem with these platforms is the challenge of protecting user privacy while running models in the cloud. Therefore Acumos gives the option to deploy the models either in the cloud or on the local hardware. To summarize the differences and similarities between Acumos and existing platforms, we present a comparison in Table I[5].

TABLE I
COMPARISON BETWEEN DIFFERENT PLATFORMS

|  | Acumos | ModelDB | OpenML | TFS | CloudML AWS ML Azure ML |
| --- | --- | --- | --- | --- | --- |
| Easy cloud deployment | Yes | No | No | Yes | Yes |
| Easy local deployment | Yes | No | No | Yes | No |
| Open source | Yes | Yes | Yes | Yes | No |
| Model sharing | Yes | Yes | Yes | No | No |
| Share model source code | Partial | Yes | No | No | No |
| Share Docker image | Yes | No | No | No | No |
| Share pre-trained model | Yes | No | No | No | No |
| Support multi-libraries | Yes | Yes | Yes | No | Partial |
| Support multi-languages | Yes | Yes | Yes | Yes | Yes |

## III. ACUMOS DESIGN AND PROCESS

The complete architectural design of Acumos is large and includes many components [8], but the scope of this paper deals with the ML-related pieces of that design, ignoring the front-end elements such as user interface, platform management, and design studio. This paper focuses specifically on how ML models are on-boarded, packaged into microservices, and shared with others.

Acumos aims to build an open ecosystem in which there are three stakeholders. The main interaction flow between these stakeholders is shown in Figure 1. A model contributor contributes models via uploading (on-boarding) the models to the platform. An end user downloads the models and uses them in their own applications. The Acumos Platform maintains the platform and enforces standards of sharing models. It is noted that a contributor can also be an end user; the platform can also be used to house models and microservices for your own use, as well as to share them with others.

The main flow is divided into three stages:
- *Uploading*: This starts with a contributor (i.e., modeler, data scientist) uploads a pre-trained local model onto

---

[1]https://www.kubeflow.org/
[2]https://cloud.google.com/ml-engine/
[3]https://aws.amazon.com/machine-learning/
[4]https://azure.microsoft.com/en-us/services/machine-learning-studio/

[5]limited to our best understanding of the platforms

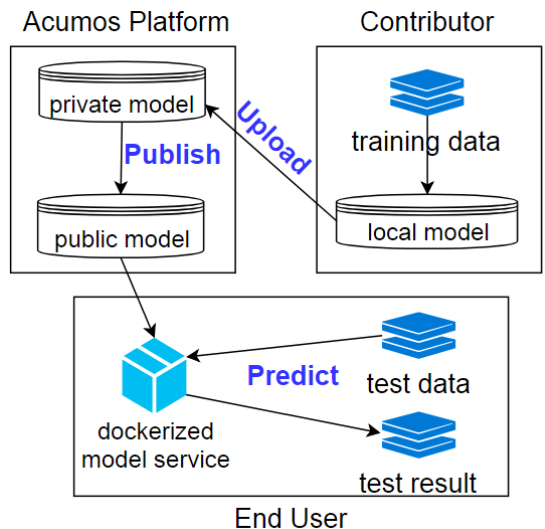

Fig. 1. Stakeholders and their relationship in the Acumos ecosystem

Acumos. A contributor is able to use a variety of existing ML libraries to build and train his/her ML model. Acumos provides Python, R and Java APIs that allow contributors to upload their models. The flexibilty to use the language and ML toolkit of choice is beneficial, because data scientists are often reluctant to change their preferred ML environment [16]. Thus a contributor can use his/her prefered environment to create their models.

- *Publishing*: Once uploaded, the models are stored in a private area, where only the contributor can access them. Then the contributor can choose when and how to share his/her model. Since the model is encapsulated and exists as something of a "black box", the contributor is required to describe the model's metadata when publishing, such as the function description, input and output format and model category. Other users can search the models based on the metadata and find the suitable ones.
- *Predicting*: The Acumos platform will pack the uploaded model as a microservice in a Docker image, which is ready to be deployed and perform its function (prediction, classification, etc.). Docker[6] provides container virtualization and it has faster speed and better agility and portability compared with virtual machines. The major reason is hardware is being virtualized to run multiple Operating Systems (OS) instances in virtual machines, but Docker offers multiple workloads running on a single instance [21]. Consumers can download and directly deploy the Dockerized service to the cloud or to any local hardware that supports Docker. Once deployed, users can send input to this running microservice and receive the its output via a RESTful API.

[6]https://www.docker.com/

## IV. ACUMOS ML COMPONENT ARCHITECTURE

The section describes the architecture of Acumos ML Onboarding component, which is shown in Figure 3 and is developed in Java[7]. We focus on the challenges and core components.

### A. Model Interfaces Unification

A feature of Acumos is its drag-and-drop design studio. In order to be able to safely reuse and compose models together, we need to understand and describe the interfaces (i.e., input and output) of models. Acumos uses Protocol Buffers[8] to offer cross-language and cross-platform support. Protocol Buffers are used for serializing the training and test data for transfer between local clients and services, and between services. Protocol Buffers supports multiple languages and they are cross-platform. The largest advantage of protocol buffers is that it is compact and efficient compared to XML-like schemes, with the ability to process large volume of data.

### B. Model Serialization

When uploading models, the Acumos API need to serialize the model locally and then deserialize it within the Acumos platform. The most challenging aspect of this is understanding the model's software dependencies. Acumos does not require users to manually provide a dependency list; instead we introspect models to infer these dependencies. In the Python client library, we traverse the object during serialization and then identify and record required libraries. Additionally, we use static analysis techniques to inspect the source code and find required modules.

### C. Microservice Generation and Deployment

The Acumos platform packs the model into a Docker image which can be deployed to an appropriate run-time environment. The Docker image is stored in the Acumos database. An end user is then able to download the Docker image and start the Dockerized model service. He/she can utilize that microservice to predict on his own application data.

The final step is to deploy a working application into a run-time system. Acumos packages solutions into Docker images which can then be deployed into any Docker environment and managed through a set of container management tools, such as Kubernetes [22]. Such image files can be easily deployed to Google Cloud, Azure, AWS, or other popular cloud services, to any corporate data center or any real-time environment as long as it supports Docker. Then the application developers can connect to the Dockerized microservice via a REST API.

## V. CASE STUDY

Sharing pre-trained models may cause problems if a modeler and an end user have different data set sources. The goal of case study A is to demonstrate Acumos works in the setting of different data sources of training and test. Furthermore, the study shows that Acumos saves lot of effort for end users

[7]source code: https://gerrit.acumos.org/r/gitweb?p=on-boarding.git
[8]https://github.com/google/protobuf

```python
from acumos.session import AcumosSession
from acumos.modeling import Model, List, create_namedtuple

import numpy as np
import pandas as pd
from sklearn.datasets import load_iris
from sklearn.ensemble import RandomForestClassifier
from sklearn.model_selection import train_test_split

#1. load the training and test data
iris = load_iris()
X = iris.data
y = iris.target
X_train, X_test, y_train, y_test = train_test_split(
    X, y, test_size=0.33, random_state=42)

#2. define the model and train the model
clf = RandomForestClassifier(random_state=0) #use either exising model or define your own model
clf.fit(X_train, y_train)                    #train the model

#3. define the Acumos Input and Ouput Type
Input = create_namedtuple('Input', [('sepal_length', List[float]),
                                    ('sepal_width', List[float]),
                                    ('petal_length', List[float]),
                                    ('petal_width', List[float])])
Output = create_namedtuple('Output', [('IrisType', List[int])])

def classify_iris(df: Input) -> Output:
    X = np.column_stack(df)
    pred = Output(IrisType = clf.predict(X))
    return pred

#4. upload the model to Acumos
iris_model = Model(classify=classify_iris)   #define Acumos model
session = AcumosSession()                    #new an Acumos session
session.push(iris_model, 'iris_model')       #upload the model
```

Fig. 2. An example of uploading a model to Acumos platform using Acumos's Python API. This is a simple predictor for the Iris dataset [20]. The Iris dataset contains 3 classes of 50 instances each, where each class refers to a type of iris plant. The features are the length and width of sepal and petal.

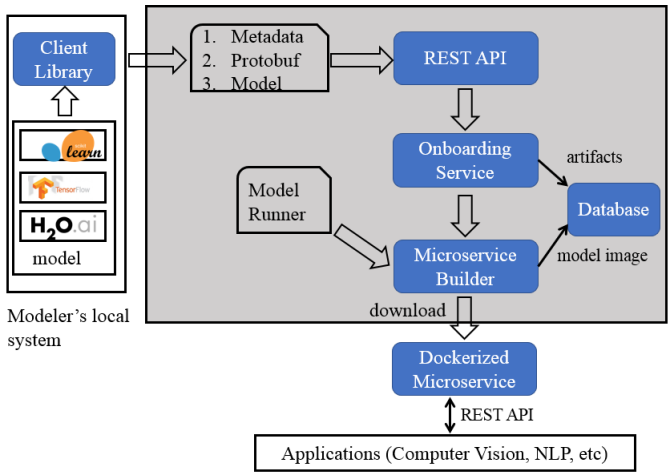

Fig. 3. The Acumos Onboarding System Architecture

to label the training data, expertise to train the model, and provides training data privacy.

The case study B is to demonstrate the microservice generated by Acumos is interoperable with any other Acumos microservice, regardless of whether it was built with any other supported toolkit. Acumos provides model-level isolation and facilitates the collaboration of different engineering teams.

A. Sentiment Analysis

Sentiment analysis is the task of identifying positive and negative opinions and emotions in texts. Here we assume there is an end user who wants to apply sentiment analysis onto her Amazon movie review dataset [23]. But she has limited knowledge on ML concepts. We will evaluate the approaches with and without Acumos. With Acumos, an end user utilizes an existing pre-trained model on Acumos. Without Acumos, she would have to train the model by herself. Below are the settings to simulate those two approaches.

*With-Acumos approach:* We train the model with the 50,000 IMDB reviews and test on the Amazon reviews. The test data is 500 random reviews from Amazon reviews [23]. This experiment setting is used to simulate the model-sharing process from a modeler to an end user via the Acumos platform.

*Without-Acumos approach:* We train the model and test the model both based on Amazon reviews. We assume that IMDB dataset is not accessible for the end user.

The training process often requires heavy computational resources which an end user typically does not possess. In addition, it may require a lot of human effort to label the data, which is also expensive. Therefore in the Without Acumos case, we gradually increase the size of training samples of Amazon reviews.

It is noted that the test data are the same for both approaches, i.e., 500 random records from Amazon reviews and we use the same model for two approaches. Inspired by previous work on sentiment analysis [24], [25], we build a simple recurrent covoluntional neural network using Keras with Tensorflow as a backend. The input to the model is a list of movie reviews, and the output is the probability of the input expressing positive emotion. Our model details are open sourced in Github[9]. The only difference of two appraoches is in the training data sets: the first approach utilizes the 50,000 IMDB reviews for training and the second approach trains the model with gradually increasing Amazon reviews.

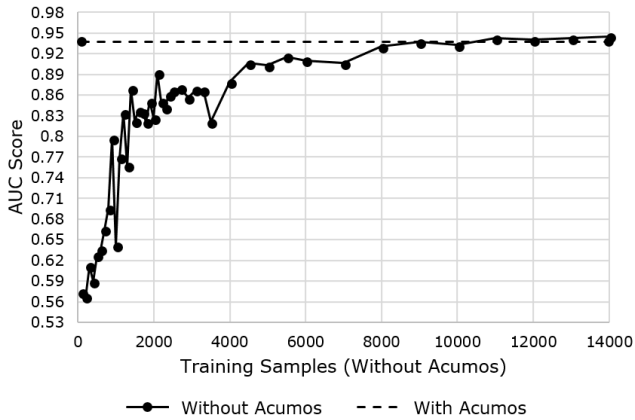

Fig. 4. Performance of Sentiment Analysis

The two approaches can be compared using the Area Under Curve (AUC) score. The AUC score is a common evaluation metric for binary classification problems, measuring the area under a receiver operating characteristic (ROC) curve. The higher the AUC score is, the better performance the model has. The result is show in Figure 4. The AUC score for Acumos approach is 0.9376, which beats the without-Acumos approach when fewer than about 10,000 training samples are used. This means that an end user would need to train on at least Amazon 10,000 reviews in order to achieve the performance level of the prepackaged model he or she could simply download from Acumos.

"With Acumos", modelers can easily pack their trained model into a microservice and share to end users who have a similar need. It relieves the user of the burdensome task of training, which involves many trial-and-error iterations, and requires expertise in ML that he or she may not have. It saves lots of effort to label the data as well. Through this experiment, we show that an end user can utilize the pre-trained models directly and with good performance, even though there is inconsistency in training data sources. In addition, Acumos helps protect data privacy, since the modeler may not wish to share the private source data (i.e., IMDB reviews here) used for training with end users.

[9]https://github.com/cheershuaizhao/acumos_ml

### B. Image Recognition and Object Detection

A hot research area in ML is object detection in images [26], such as traffic light detection, face detection, gender detection, and landmark detection [27]. An image contains rich information, and thus there are numerous detection tasks we could potentially apply to it. However, the training of ML on images is expensive and time-consuming. It involves a large amount of human effort in labelling, and requires high-performance computing resources for training models. Existing techniques of Multiple-Task Learning try to learn multiple tasks together, but they requires the tasks to have some similarities as well as careful parameter tuning by modelers [11].

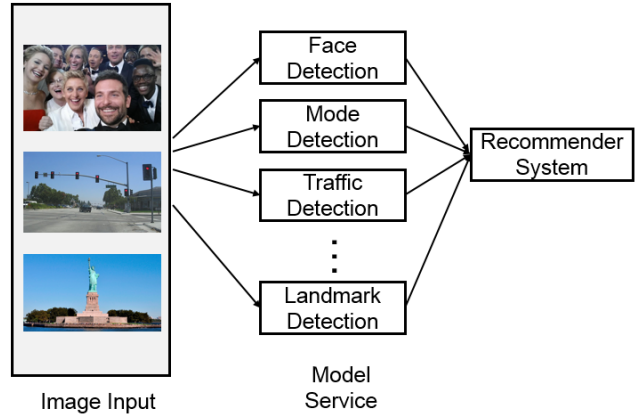

Fig. 5. The Combination of Acumos Model Services

As shown in Figure 5, an application developer can easily combine different object detection microservices and she can append additional actions after the model services, such as adding a rule-based recommendation function. These microservices can be consumed by simply making calls to the API functions similar to any other REST services. The appended recommendation system could also be another microservice, which takes the output of object detection microservices as its own input and generates the recommended item.

The model services are pre-trained by model experts on high-quality data, which guarantees their good performance. This mechanism of model-level isolation facilitates reuse when coupled with the ability to compose existing models into complex composite solutions. Models can be easily composed in parallel or sequentially. In addition, modelers are not restricted to use the same set of toolkits to develop their models. They can pick their favorite suitable tools for each object detection task.

## VI. LIMITATION

Most of these limitations follow directly from the design of Acumos which treats ML models as black boxes and focuses on the sharing process. Thus Acumos does not optimize the execution of models within their respective ML frameworks. It is the modelers' responsibility to achieve good model performance. The black-box approach also incurs some communication overhead and duplication of function over a single-model pipelined approach. Finally, Acumos in its current state lacks facilities to validate and test models on the platform itself.

## VII. CONCLUSION

This paper introduced Acumos Platform, a system for packaging and sharing ML models. Acumos makes ML models accessible to a wide audience by creating an extensible marketplace for models. Using Acumos, data scientists can use their familiar ML toolkits and libraries to create models, and then share them with ordinary developers who are not ML experts.

We addressed the challenges of integrating ML models into application development and model sharing. The proposed architecture supports the sharing of pretrained models across different ML libraries and run-time environments. As illustrated by the case studies, Acumos ML provides model-level isolation and focuses on model reusability, rather than the model development process. Acumos offers an open marketplace where the accessing ML models is secure and convenient. Currently the marketplace is designed for model sharing and distribution free but it retains the potential of pricing and gaining revenue especially for models with high quality.

In the future, we hope to create standards or guidelines for model input/output format definition as well as data mapping rules. Also, we propose to add model validation procedures when modelers upload models.

## ACKNOWLEDGMENT

The authors would like to thank the entire development team from AT&T and Tech Mahindra, who collaborated to develop the initial Acumos platform. We thank Mazin E Gilbert for his invaluable support of the Acumos project.